\documentclass{article}

\usepackage[final]{neurips_2024}

\usepackage[utf8]{inputenc} % allow utf-8 input
\usepackage[T1]{fontenc}    % use 8-bit T1 fonts
\usepackage{hyperref}       % hyperlinks
\usepackage{url}            % simple URL typesetting
\usepackage{booktabs}       % professional-quality tables
\usepackage{amsfonts}       % blackboard math symbols
\usepackage{nicefrac}       % compact symbols for 1/2, etc.
\usepackage{microtype}      % microtypography
\usepackage{xcolor}         % colors
\usepackage{amsmath, amsfonts}
\usepackage[noabbrev]{cleveref}
\usepackage{tabularx, caption, boldline, adjustbox}
\usepackage{graphicx}
\usepackage{array}
\usepackage{colortbl}
\definecolor{Gray}{gray}{0.9} % Define a custom gray color
\usepackage{multirow}
\usepackage{booktabs}
\usepackage{tcolorbox}
\tcbuselibrary{breakable}
\usepackage{soul} % For highlighting text
\usepackage{xcolor} % To define custom highlight colors
\definecolor{lightyellow}{rgb}{1,1,0.8}
\definecolor{lightred}{rgb}{1.0, 0.8, 0.8}
\definecolor{lightgreen}{rgb}{0.8,1,0.8}
\usepackage{framed}
\definecolor{promptbg}{RGB}{240,240,240}
\definecolor{promptframe}{RGB}{200,200,200}

\usepackage{caption}
\title{Positive Experience Reflection for Agents in Interactive Text Environments}

\author{%
  Philip Lippmann \hspace{16pt} Matthijs T.J.~Spaan \hspace{16pt}  Jie Yang   
  \vspace{4pt}\\
  Delft University of Technology\\
  \small\texttt{\{p.lippmann, m.t.j.spaan, j.yang-3\}@tudelft.nl} \\
}

\begin{document}

\maketitle

\begin{abstract}
Intelligent agents designed for interactive environments face significant challenges in text-based games, a domain that demands complex reasoning and adaptability.
While agents based on large language models (LLMs) using self-reflection have shown promise, they struggle when initially successful and exhibit reduced effectiveness when using smaller LLMs. 
We introduce Sweet\&Sour, a novel approach that addresses these limitations in existing reflection methods by incorporating positive experiences and managed memory to enrich the context available to the agent at decision time.
Our comprehensive analysis spans both closed- and open-source LLMs and demonstrates the effectiveness of Sweet\&Sour in improving agent performance, particularly in scenarios where previous approaches fall short.
%
%This work contributes to advancing the capabilities of LLM-based agents in complex, dynamic environments such as TBGs.
\end{abstract}

\section{Introduction}
Intelligent agents, designed to interact with and make decisions in dynamic environments, have become a central focus in AI research, with text-based games (TBGs) emerging as a particularly challenging domain for evaluating these agents' reasoning, adaptability, and learning abilities~\citep{cote18textworld, wang2022scienceworld}. 
Originally popular in the 1970s as text adventure games,\footnote{Try it yourself: \url{https://www.microsoft.com/en-us/research/project/textworld/try-it/}} TBGs present players with textual descriptions of environments, requiring them to input natural language commands to achieve objectives~\citep{hausknecht2020interactive}. 
For instance, determining if a metal fork is conductive involves locating the fork, assembling a circuit, and analyzing the result. 
Navigating TBGs demands that agents exhibit a combination of abilities, including planning, memory retention, spatial reasoning, and common sense knowledge~\cite{wang2023interactivenaturallanguageprocessing}.

Previously, deep reinforcement learning and behavior cloning were the primary approaches to develop agents to play TBGs~\cite{ammanabrolu2019playing, yao2020keep}.
However, recent research shows that agents based on pretrained large language models (LLMs) are more effective at navigating TBGs~\cite{Lin2023SwiftSageAG}. 
A key factor in their success is the integration of internal \emph{reflection} to improve planning~\cite{xi2023risepotentiallargelanguage, huang2024understandingplanningllmagents, hu2024surveylargelanguagemodelbased}.

Self-reflection, closely related to self-refinement, is a form of reasoning that occurs after receiving binary or scalar feedback from the environment~\cite{madaan2023selfrefine}.
In this process, the LLM reviews its actions and their outcomes, considering what went wrong and potential ways to improve~\cite{wang2024describeexplainplanselect}. 
By iteratively adjusting its strategy based on verbal reinforcement, conveyed through textual feedback, the agent refines its planning for subsequent attempts~\cite{shinn2023reflexion}. 
However, reflection also has several limitations, including 1) underwhelming performance when agents are correct initially~\cite{li2024hindsight2020testinglimits}, 2) significantly worse efficacy when using smaller LLMs~\cite{Lin2023SwiftSageAG}, and 3) dependence on external feedback~\cite{zhang2024selfcontrastbetterreflectioninconsistent}.

\noindent\textbf{Our Contributions}
In this work we conduct a comprehensive analysis of LLM-based agents employing reflection approaches in TBGs and evaluate their performance across closed- and open-source LLMs.
To address the limitations of poor performance when agents are initially successful and the diminished efficacy of smaller LLMs, we propose \emph{Sweet\&Sour} to leverage positive experiences and managed memory to create a richer context for self-reflection.
Our findings demonstrate that our method improves the performance of agents using reflection, particularly in scenarios where they previously struggled, enabling more robust and generalizable learning across tasks and model sizes.

\begin{figure*}[t]
    \centering
    \includegraphics[width=\textwidth]{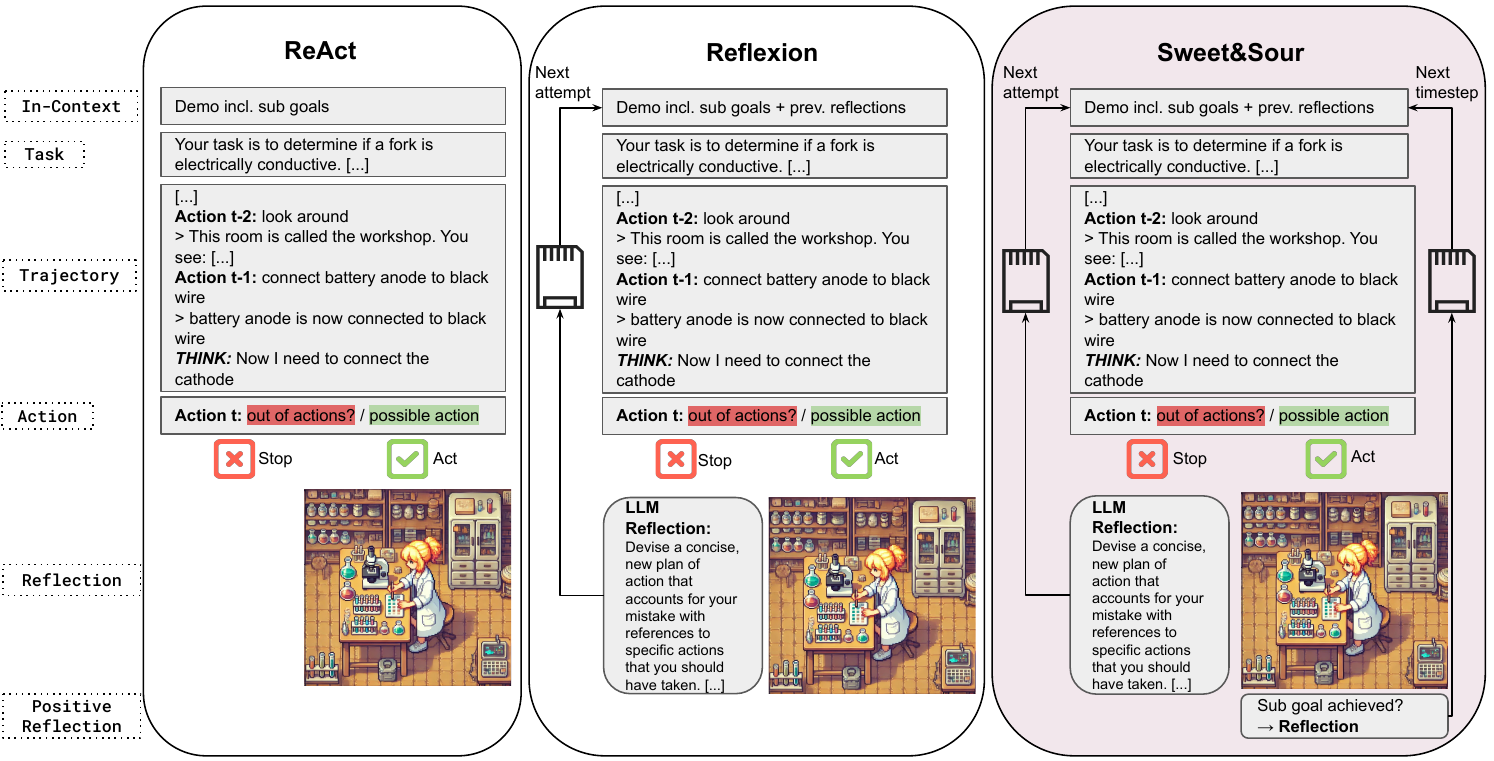}
    \caption{Comparison of used prompting methods to play ScienceWorld. ReAct introduces a \texttt{THINK} action to explicitly reason regarding the next step. Reflexion leverages self-reflection across attempts to learn from unsuccessful tries and stores these in memory. Sweet\&Sour not only performs self-reflection after failures but also after each completed sub goal, making its reflection instantly available.}
    \label{fig:method}
\end{figure*}
\section{Methodology}

\noindent\textbf{Background}
Assuming an LLM behaving as an actor model as part of our agent -- \emph{i.e.} generating actions based on the current state and policy, analogous to traditional policy-based RL setups -- we sample an action $a_t$ from the current policy $\pi_{\theta}$ at time $t$ and receive an observation from the environment $o_t$.
When a game begins, the agent makes its first observation $o_{0}$ at time step $t=0$.
This first observation differs from subsequent ones, as it consists of the goal description $d$, as well as an analysis of the starting room (\emph{i.e.} the output of the ``look around'' command).
Subsequently, the agent can perform an action $a_{t}$ at each time step and receives a corresponding observation $o_{t}$ from the environment.
The agent has an inventory $i_{t}$ in which to store items.
Each task consists of a number of sub tasks (such as finding a key object), the completion of which grants the agent a sparse reward, which adds to its current reward $r_t$.
The game continues until the agent has achieved the goal outlined in $d$ and receives the full reward as final score, or the maximum number of steps -- which we set to 150 -- is reached, in which case $r_t$ will become the final score.
A detailed problem formulation and assumptions we make are given in \cref{sec:background}.

\noindent\textbf{Self-Reflection}
Reflection occurs in addition to the acting LLM. 
Here, the agent reviews the $a_t$ and $o_t$ associated with previous unsuccessful attempts to verbalize the reason for failure.
This process typically involves maintaining a persistent history of insights gained across attempts, which the LLM uses as additional context for its reflections to improve future decision making for the next attempt~\cite{shinn2023reflexion}.
However, since other self-reflection methods focus on learning from failures~\cite{renze2024selfreflectionllmagentseffects, zhang2024proagentbuildingproactivecooperative, zhang2024agentprolearningevolvepolicylevel, huang2024recommenderaiagentintegrating, yao2024retroformer}, they overlook the importance of reinforcing successful behaviors in a similar way.

\noindent\textbf{Sweet\&Sour}
To address the limitations of existing reflection methods, we enhance self-reflection by incorporating both positive (\emph{sweet}) and negative (\emph{sour}) experiences into the reflection process.
This enables the agent to learn effectively from successful actions by reinforcing strategies that lead to positive outcomes while still learning from failures.
When the current policy is achieving rewards, we query the agent to extrapolate from it, encouraging the agent to verbalize what made its current policy successful and what can be generalized from this.
This is visualized in \cref{fig:method} and an example of this is shown in \cref{sec:app-llm}.
Our method is broadly applicable to agents in interactive text environments with feedback using self-reflection, including those that build additional complexity on top of the core reflection loop, such as grounding~\cite{Lin2023SwiftSageAG} or gradient learning~\cite{yao2024retroformer}.
We leave the study of these additional use cases for future work.

Previous works store their reflections gained from unsuccessful attempts in something akin to a long-term memory to make them available to the agent across attempts~\cite{huang2024recommenderaiagentintegrating, shinn2023reflexion}.
Instead, to complement Sweet\&Sour, we propose a \emph{managed memory} approach to store and retrieve relevant reflections. 
This is implemented using a dual-buffer structure, where experiences are stored in two categories: short-term memory and long-term memory, based on their outcome (success or failure) and recency.
Initially, if sub goals are reached, short-term memories of the made reflections are stored in a temporary buffer.
Each short-term memory consists of a tuple $(\text{reflection}_t, o_t, a_t, r_t)$.
Once a task is completed or an attempt ends, all short-term memories are moved to long-term memory.
Failed attempts' reflections are immediately added to long-term memory for the next attempt and short-term memory collection ends.

\section{Experiments}

\noindent\textbf{Data \& Environment}
We use the ScienceWorld benchmark~\cite{wang2022scienceworld}, which provides a versatile setting for evaluating agents in science experiment tasks across 10 interconnected locations, such as a greenhouse and a workshop, with over 200 objects and 25 action templates, generating a vast and dynamic search space.
We use the test set for our evaluation, which provides up to 10 variations of each of the 30 distinct tasks.
These interactive tasks cover various topics, including chemistry and electricity, and have an average optimal decision depth of 50 steps.
An example task is shown in \cref{sec:example-task}.
For details of all tasks and the environment, we refer to~\cite{wang2022scienceworld}.
We elect to use ScienceWorld instead of previous interactive text environment benchmarks such as TWC~\cite{Murugesan2021twc} and ALFWorld~\cite{ALFWorld20}, due to their relative simplicity for current LLM-based agents.
We measure performance using the success score, which is always between 0 and 100.
Completing a task implies completing every sub task, receiving the full reward, and thus a score of 100.

\noindent\textbf{Baselines}
CALM~\cite{yao2020keep} is a reranking method that integrates a deep reinforced relevance network (DRRN)~\cite{he-etal-2016-deep} with a causal language model fine-tuned using oracle trajectories. 
The causal language model acquires task- and environment-specific knowledge through imitation learning, while DRRN trains a policy network to rerank the language model's predictions.
We use ReAct~\cite{yao2023reactsynergizingreasoningacting} as our baseline LLM-based agent.
ReAct composes useful information at each time step by reasoning over the current context (\emph{e.g.} decomposing task or common sense knowledge query) and carries it forward to the context of the following time step.
This reasoning action does not affect the environment and may be considered few-shot in-context learning.
To contextualize our work, we compare our method against Reflexion~\cite{shinn2023reflexion}, an agent built on ReAct that employs a self-reflection mechanism to iteratively improve its performance across rounds upon encountering failure based on feedback from the environment.
As such, it runs over up to four rounds as it builds up its long-term memory.
For all agents, we evaluate their performance using LLMs of different sizes and complexities to assess the performance of each method across varying computational resources.
In descending order of parameter count, we select GPT-4o (\texttt{gpt-4o-2024-08-06})\cite{openai2023gpt4}, Mistral Large 2 (\texttt{mistral-large-2407}), and Llama 3.1 8B (\texttt{llama-3.1-8b-instruct})\cite{dubey2024llama3herdmodels}, accessing each through its respective APIs.

%\section{Results}
\begin{table*}[t]
\caption{Results on the ScienceWorld benchmark. For each method, we use GPT-4o (GPT), Mistral Large 2 (ML2), and Llama 8B (L8B). Each value is an average of across all task variations.}
\centering
\begin{adjustbox}{width=\textwidth}
\begin{tabular}{lcccccccccccc}
\toprule
\textbf{Task} & \multicolumn{1}{c}{\textbf{CALM}} & \multicolumn{3}{c}{\textbf{ReAct}} & \multicolumn{3}{c}{\textbf{Reflexion}} & \multicolumn{3}{c}{\textbf{Sweet\&Sour (ours)}}\\
\cmidrule(lr){2-2} \cmidrule(lr){3-5} \cmidrule(lr){6-8} \cmidrule(lr){8-11}
 & CALM & L8B & ML2 & GPT & L8B & ML2 & GPT & L8B & ML2 & GPT\\
\midrule
\textbf{1-1} (Boil) & 0.0 & 0.0 & 0.0 & 3.8 & 0.0 & 0.0 & 5.1 & 0.0 & 7.2 & 9.6 \\
\textbf{1-2} (Melt) & 0.0 & 8.4 & 10.3 & 11.8 & 0.0 & 0.0 & 10.0 & 11.4 & 12.1 & 12.8 \\
\textbf{1-3} (Freeze) & 0.0 & 1.5 & 0.0 & 8.1 & 0.0 & 2.3 & 8.3 & 2.4 & 3.1 & 8.9 \\
\textbf{1-4} (Change state) & 0.0 & 1.0 & 4.7 & 10.0 & 0.0 & 0.0 & 4.2 & 1.7 & 2.9 & 9.2 \\
\textbf{2-1} (Thermometer) & 1.0 & 5.1 & 7.8 & 7.7 & 3.4 & 4.2 & 7.6 & 7.8 & 9.7 & 10.9 \\
\textbf{2-2} (Melting) & 1.0 & 6.7 & 6.3 & 5.9 & 3.3 & 3.3 & 26.2 & 7.9 & 36.8 & 46.0 \\
\textbf{2-3} (Melting) & 5.0 & 9.1 & 11.8 & 23.4 & 13.2 & 14.7 & 22.6 & 15.2 & 29.0 & 38.3 \\
\textbf{3-1} (Power 1) & 7.0 & 18.8 & 24.6 & 57.2 & 21.2 & 51.5 & 78.4 & 28.6 & 75.4 & 81.1 \\
\textbf{3-2} (Power 2) & 2.0 & 10.2 & 24.7 & 55.6 & 9.5 & 11.9 & 24.7 & 23.3 & 44.5 & 58.0 \\
\textbf{3-3} (Conductivity 1) & 2.0 & 52.4 & 51.7 & 73.0 & 9.2 & 25.8 & 72.1 & 59.1 & 69.2 & 75.7 \\
\textbf{3-4} (Conductivity 2) & 10.0 & 54.2 & 64.9 & 89.7 & 35.4 & 41.6 & 75.1 & 62.7 & 60.3 & 67.3 \\
\textbf{4-1} (Find 1) & 54.0 & 17.3 & 18.7 & 27.5 & 44.6 & 48.1 & 62.3 & 41.7 & 71.7 & 74.2 \\
\textbf{4-2} (Find 2) & 10.0 & 69.1 & 71.6 & 80.3 & 68.4 & 75.7 & 87.3 & 76.8 & 100.0 & 100.0 \\
\textbf{4-3} (Find 3) & 8.0 & 21.3 & 42.8 & 47.7 & 18.4 & 16.5 & 17.3 & 20.9 & 21.5 & 34.3 \\
\textbf{4-4} (Find 4) & 2.0 & 15.7 & 15.2 & 19.3 & 39.6 & 46.6 & 100.0 & 55.1 & 87.8 & 100.0 \\
\textbf{5-1} (Grow plant) & 4.0 & 10.8 & 10.8 & 10.0 & 7.2 & 7.2 & 7.9 & 14.2 & 14.6 & 17.4 \\
\textbf{5-2} (Grow fruit) & 3.0 & 18.1 & 18.5 & 19.2 & 30.8 & 51.4 & 34.6 & 51.5 & 55.6 & 60.2 \\
\textbf{6-1} (Chemistry 1) & 6.0 & 37.8 & 42.9 & 58.6 & 27.1 & 29.7 & 70.2 & 37.9 & 61.1 & 70.2 \\
\textbf{6-2} (Chemistry 2) & 3.0 & 25.0 & 27.1 & 50.6 & 14.4 & 28.0 & 69.8 & 27.2 & 51.9 & 83.1 \\
\textbf{6-3} (Chemistry 3) & 6.0 & 14.4 & 17.5 & 39.7 & 38.9 & 31.1 & 16.7 & 45.3 & 53.7 & 61.5 \\
\textbf{7-1} (Lifespan 1) & 10.0 & 37.0 & 41.7 & 60.0 & 75.0 & 75.0 & 100.0 & 75.0 & 88.2 & 100.0 \\
\textbf{7-2} (Lifespan 2) & 4.0 & 50.5 & 50.7 & 67.5 & 60.0 & 71.9 & 81.4 & 70.5 & 77.0 & 80.0 \\
\textbf{7-3} (Lifespan 3) & 4.0 & 33.7 & 38.2 & 50.0 & 29.5 & 33.7 & 75.0 & 51.1 & 54.2 & 84.6 \\
\textbf{8-1} (Identify life 1) & 0.0 & 5.1 & 18.9 & 25.3 & 1.7 & 1.7 & 3.4 & 11.1 & 10.3 & 14.2 \\
\textbf{8-2} (Identify life 2) & 0.0 & 6.4 & 7.4 & 8.0 & 7.4 & 8.0 & 8.0 & 5.0 & 7.4 & 7.4 \\
\textbf{9-1} (Measure angle) & 0.0 & 28.5 & 33.0 & 42.5 & 56.9 & 55.1 & 57.1 & 68.4 & 70.3 & 75.0 \\
\textbf{9-2} (Friction 1) & 3.0 & 14.5 & 22.6 & 43.1 & 23.4 & 29.3 & 100.0 & 33.3 & 36.7 & 62.0 \\
\textbf{9-3} (Friction 2) & 2.0 & 2.9 & 14.5 & 42.8 & 1.3 & 33.6 & 59.6 & 7.2 & 51.9 & 63.1 \\
\textbf{10-1} (Genetics 1) & 2.0 & 25.7 & 27.3 & 26.4 & 5.6 & 9.8 & 50.4 & 38.9 & 48.6 & 78.8 \\
\textbf{10-2} (Genetics 2) & 2.0 & 13.2 & 19.1 & 17.2 & 6.2 & 21.5 & 22.7 & 23.6 & 24.0 & 54.8 \\
\midrule
\textbf{Average} & \textbf{5.07} &\textbf{20.5} & \textbf{24.8} & \textbf{36.0} & \textbf{21.7} & \textbf{27.6} & \textbf{45.3} & \textbf{32.5} & \textbf{44.6} & \textbf{54.6} \\
\bottomrule
\end{tabular}
\end{adjustbox}
\label{tab:results100}
\end{table*}

\noindent\textbf{Main Results}
The results are shown in \cref{tab:results100}. 
We find that Sweet\&Sour outperforms the baseline methods across all LLMs, setting the highest average score at 54.6 using GPT-4o.
The performance gap between Sweet\&Sour and the other methods widens for smaller models with a lower parameter count.
For instance, it achieves 44.6 compared to Reflexion's 27.6 on Mistral Large 2, and 32.5 compared to 21.7 on Llama 8B -- indicating that our method is more suitable for scenarios with limited computational resources.
When we modify our method to only sample from failures, performance drops significantly to a level similar to Reflexion -- scores decrease to 24.6, 31.1, and 44.9 for Llama 8B, Mistral Large 2, and GPT-4o, respectively.
As such, incorporating positive experiences indeed leads to better reflections, mimicking how humans learn from both positive and negative experiences, resulting in improved decision-making.

\noindent\textbf{Anti-Tilt}
In highly challenging tasks, such as 1-1 and 8-2, all methods tend to struggle, while in simpler tasks, most methods succeed based on the LLM’s inherent capabilities alone. 
However, medium-difficulty tasks, such as 3-2 and 3-3, reveal a critical performance gap between our method and previous approaches.
We note that this occurs because traditional methods fail to reflect on early successes, leaving them with less contextual understanding to carry momentum forward, leading to a sustained performance decline or ``tilt.'' 
By contrast, we theorize that Sweet\&Sour's reflection on both successes and failures provides a richer context, enabling it to sustain and build upon initial successes.
For instance, on task 3-2, Sweet\&Sour achieves a score of 68.0 with GPT-4o, significantly outperforming Reflexion’s 24.7 while beating ReAct’s 55.6 by a smaller margin, indicating that Reflexion, although it is overall more capable, got stuck despite its self-reflection.

\section{Conclusion}
In an attempt to improve agent performance and reduce sensitivity to the LLM used, our study embarks on an analysis of reflection mechanisms for LLM-based agents. 
Existing work focuses on learning from failures only.
Although these methods improve planning by analyzing past mistakes, they struggle when agents are initially successful and when using smaller LLMs. 
The primary contribution of our work is Sweet\&Sour, a novel reflection technique for LLM-based agents in TBGs that leverages positive experiences to improve agent self-reflection. 
Our comprehensive analysis demonstrates the effectiveness of Sweet\&Sour in enhancing agent adaptability and reasoning, particularly in challenging situations where previous approaches struggled. 
%
%These findings not only advance the field of intelligent agents in text-based environments but also suggest potential applications in broader domains requiring complex decision making.

\noindent\textbf{Limitations}
Despite promising results, our work has limitations. 
LLMs do not provide guarantees regarding their reasoning capabilities.
Additionally, our evaluation is conducted using a single environment, which, while comprehensive, does not cover all types of interactive scenarios. 
We leave the exploration of additional environments to future work.

\section*{Acknowledgments}
This work was supported by an Oracle
for Research Grant Award, as well as SURF Grant EINF-8535.

\bibliographystyle{plainnat}
\bibliography{custom}

\appendix
\clearpage
\section{Background}
\label{sec:background}
One may consider every TBG to be a partially observable Markov decision process (POMDP)~\citep{spaan2012partially} where the environment state is never observed directly.
This POMDP may be formalized as $\langle S, T, A, \Omega, R, \gamma\rangle$, where $\gamma \in [0,1]$ denotes the reward discount factor.
$S$ denotes the set of states $s$ that contain the internal information of the game -- such as objects found throughout the game or the player's location -- not all of which may be visible to the agent at any given time.
$A$ denotes the action space made up of individual text actions $a$ issued by the player.
$\Omega$ denotes the observation function.
Further, $o \in O$ denotes the observations made by the player.
The observation $o_{t}$ of the agent at time $t$ depends on the current state $s_{t}$, as well as the previous action $a_{t-1}$, which may be formalized as $\Omega\left(o_{t} \mid s_{t}, a_{t-1}\right)$.
Seeing as the agent can only observe and interact with the environment of a TBG via natural language, each observation is composed of a sequence of tokens $o_{t} = \left\{\hat{o}^{1}_{t}, \cdots, \hat{o}^{N}_{t} \right\}$, as are their actions $a_{t} = \left\{ \hat{a}^{1}_{t}, \cdots, \hat{a}^{M}_{t} \right\}$.

%The environment transitions to its next state $s_{t+1}$, once an admissible action is performed by the agent, $T\left(s_{t}, a_{t}\right) = s_{t+1}$, where $T$ denotes the state transition function.
%
%An action may be considered admissible at the current state if it changes the underlying game state $T\left(s_{t}, a_{t}\right) \neq s_{t}$.
%
%We make the assumption that the player can choose an action from a set consisting only of admissible actions.
%
%The admissible action set is bound to be significant for quests with a sufficiently large branching factor. 
%
%While admissible, many potential actions are bound to be suboptimal and as such reducing the size of this set without removing the action candidates that lead to quest completion should improve sample efficiency.
%%%%
%
In the context of TBGs, an action $a_{t}$ is considered admissible at a state $s_{t}$ if it is capable of changing the game's state, \emph{i.e.}, if it can lead to a transition to a new state $s_{t+1}$ that is different from the current state $s_{t}$. 
The environment's state transition is modeled through a probabilistic function $T\left(s_{t+1} \mid s_{t}, a_{t}\right)$. 
Traditionally, admissible actions in state $s_{t}$ could deterministically lead to a new state $s_{t+1}$. 
However, we use a more general approach where all actions, whether admissible or not, are included in the state transition function. 
Non-admissible actions, which do not lead to a change in the game's state, result in a transition back to the original state $s_{t}$ with probability 1. 
In contrast, admissible actions lead to different states with their own probability. 
The admissible action set is bound to be significant for quests with a sufficiently large branching factor. 
While admissible, many action candidates are bound to be suboptimal.

The reward $r$ received by the agent -- the discounted sum of which, $\mathbb{E}\left[\sum_{t}\gamma^{t}r_{t}\right]$, it aims to maximize -- are obtained by $r_{t} = R\left(s_{t}, a_{t}\right)$.
In practice, TBGs typically provide sparse positive rewards for completing subgoals and advancing towards completing the game.

A TBG's interpreter can accept any text sequence, but will only recognize text that follows a certain structure.
Typically, commands take the form of either a single keyword (such as ``look'') or a combination involving verbs, objects, and occasionally prepositions.
Previous works on TBGs made the assumption that we have access to a predefined set of all admissible actions at each game state and must select the correct action to progress~\citep{narasimhan2015language, he2016deep}.
This is consistent with how some text adventure games are played in the real world~\citep{tao2018solving}.
In line with more recent works~\cite{Lin2023SwiftSageAG}, we make the assumption that we have access to a number of action templates (\emph{e.g.} \texttt{connect A to B}, \texttt{pick up C}) and subsequently ask the agent to generate the actions as tokens using these templates.
This leads to a more challenging action generation process for the agent.
\section{LLM Reflection}
\label{sec:app-llm}

After successfully completing a sub task, the agent is prompted to reflect on its most recent observations and identify the key factors that contributed to its success. 
This reflection process encourages the agent to verbalize the steps and strategies that led to the desired outcome, helping it create a plan that can be stored in managed memory for future use. 
A similar procedure is applied in cases of failure. 
When a task is unsuccessful, the agent is instructed to reflect on alternative actions it could have taken and to devise a revised plan for the next attempt, ensuring continuous learning and improvement. 
This plan is also stored in memory.

An example trajectory of a ScienceWorld task where the agent reflects on positive and negative experiences is shown in Figure 2.
At the end of this trajectory, we show for a single command what a reflection would look like for a successful or unsuccessful choice, in this case picking up an animal.
For the unsuccessful case (red), where the agent does not pick up an animal, we reflect on whether another choice would have constituted an animal and thus resulted in a reward.
For the successful case where a sub goal is reached (yellow), we reflect on what made the current actions successful and subsequently commit this to memory.

\begin{tcolorbox}[colback=gray!10!white, colframe=gray!50!black, breakable, width=\textwidth, title=ScienceWorld Task 8-1, boxrule=0.5mm, arc=2mm, outer arc=2mm]

    \textbf{Task Description:} \\
    Your task is to find a(n) animal. First, focus on the thing. Then, move it to the red box in the kitchen.

    \vspace{10pt}

    \texttt{> look around} \\
    This room is called the hallway. In it, you see:
    \begin{itemize}
        \item a substance called air
        \item a picture
        \item the agent
    \end{itemize}
    You also see:
    \begin{itemize}
        \item A door to the greenhouse (open)
        \item A door to the living room (open)
        \item A door to the art studio (open)
        \item A door to the kitchen (open)
        \item A door to the bedroom (open)
        \item A door to the workshop (open)
    \end{itemize}

    \texttt{> open door to greenhouse} \\
    The door is already open.

    \texttt{> go to greenhouse} \\
    You move to the greenhouse.

    \texttt{> open door to outside} \\
    The door is already open.

    \texttt{> go to outside} \\
    You move to the outside.

    \texttt{> look around} \\
    This outside location is called the outside. Here, you see:
    \begin{itemize}
        \item a substance called air
        \item an axe
        \item a substance called wood
        \item the ground
        \item a fire pit (containing nothing)
        \item a fountain (containing a substance called water)
        \item a butterfly egg
        \item a blue jay egg
        \item a dove egg
        \item the agent
    \end{itemize}
    You also see:
    \begin{itemize}
        \item A door to the kitchen (open)
        \item A door to the foundry (open)
        \item A door to the greenhouse (open)
    \end{itemize}

    \textbf{Case \#1: Failure}
    
    \texttt{> focus on axe} \\
    You focus on the axe.

    \texttt{\sethlcolor{lightred}\hl{> pick up axe}} \\
    \sethlcolor{lightred}\hl{You move the axe to the inventory.}

    \emph{Reflection(I failed because I assumed eggs are not animals. My plan was to go outside to find animals. I tried to pick up an axe. An alternative course of action would have been to go outside and pick up an egg, as it might be considered an animal.)}\\
    $\rightarrow$ Managed Memory

    \vspace{5pt}
    \tcblower
    \vspace{5pt}
    \textbf{Case \#2: Success}

    \texttt{> focus on dove egg} \\
    You focus on the dove egg.

    \texttt{\sethlcolor{lightyellow}\hl{> pick up dove egg}} \\
    \sethlcolor{lightyellow}\hl{You move the dove egg to the inventory.}

    \emph{Reflection(Picking up an egg was the correct choice. Even though the animal was in an egg, it is still considered an animal. My plan was to go outside and find an animal in the wild. Items such as an axe, wood, or landscape elements are not related to animals.)}\\
    $\rightarrow$ Managed Memory\\

\end{tcolorbox}
\begin{center}
    Figure 2: Example of reflection in the 2-1 Task, where the agent needs to find something.
\end{center}

\section{Example ScienceWorld Task}
\label{sec:example-task}
In this section, we provide a successfully completed task, a variation of Task 1-1, which concerns itself with boiling a substance, in this case water, to change its state.
Once the agent has viewed its surroundings and moved to the kitchen, it collects the necessary items and begins its experiment by boiling the water in a pot on the stove.
Finally, the agent examines steam and completes the task (highlighted in green).
The example is truncated to improve readability.
The trajectory is shown in Figure 3.

%\sethlcolor{lightyellow}\hl{}
\begin{tcolorbox}[colback=gray!10!white, colframe=gray!50!black, breakable, width=\textwidth, title=ScienceWorld Task 1-1, boxrule=0.5mm, arc=2mm, outer arc=2mm]

    \textbf{Task Description:} \\
    Your task is to boil water. For compounds without a boiling point, combusting the substance is also acceptable. First, focus on the substance. Then, take actions that will cause it to change its state of matter.

    \vspace{10pt}

    \texttt{> look around} \\
    This room is called the hallway. In it, you see:
    \begin{itemize}
        \item a picture
        \item a substance called air
        \item the agent
        \item A door to the green house (open)
        \item A door to the living room (open)
        \item A door to the art studio (open)
        \item A door to the kitchen (open)
        \item A door to the bedroom (open)
        \item A door to the workshop (open)
    \end{itemize}

    \texttt{> open door to kitchen} \\
    The door is already open.

    \texttt{> go to kitchen} \\
    You move to the kitchen.

    \texttt{> look around} \\
    This room is called the kitchen. In it, you see:
    \begin{itemize}
        \item a substance called soap
        \item a painting
        \item a thermometer reading 10°C
        \item a counter with various items (banana, potato, red apple, orange)
        \item a sink (off, empty)
        \item a table with a glass cup (empty)
        \item a chair (empty)
        \item a freezer (closed)
        \item a lighter
        \item a stopwatch (deactivated)
        \item a fridge (closed)
        \item a cupboard (closed)
        \item a stove (turned off)
        \item a glass jar containing sodium chloride
        \item various other items (air, the agent, etc.)
    \end{itemize}

    \texttt{> pick up thermometer} \\
    You move the thermometer to the inventory.

    \texttt{> open cupboard} \\
    The cupboard is now open.

    \texttt{> pick up metal pot} \\
    You move the metal pot to the inventory.
    
    \texttt{> move metal pot to sink} \\
    You move the metal pot to the sink.

    \texttt{> activate sink} \\
    The sink is now activated.

    \texttt{> deactivate sink} \\
    The sink is now deactivated.

    \texttt{> pick up metal pot} \\
    You move the metal pot to the inventory.

    \texttt{> focus on substance in metal pot} \\
    You focus on the water.

    \texttt{> move metal pot to stove} \\
    You move the metal pot to the stove.

    \texttt{> activate stove} \\
    The stove is now activated.

    \texttt{> use thermometer in inventory on substance in metal pot} \\
    The thermometer measures a temperature of 13 degrees Celsius.

    \texttt{> examine substance in metal pot} \\
    A substance called water.

    \texttt{> use thermometer in inventory on substance in metal pot} \\
    The thermometer measures a temperature of 28 degrees Celsius.

    \texttt{> examine substance in metal pot} \\
    A substance called water.

    \texttt{> use thermometer in inventory on substance in metal pot} \\
    The thermometer measures a temperature of 58 degrees Celsius.

    \texttt{> examine substance in metal pot} \\
    A substance called water.

    \texttt{> use thermometer in inventory on substance in metal pot} \\
    The thermometer measures a temperature of 78 degrees Celsius.

    \texttt{> examine substance in metal pot} \\
    A substance called water.

    \texttt{> use thermometer in inventory on substance in metal pot} \\
    The thermometer measures a temperature of 98 degrees Celsius.

    \texttt{\sethlcolor{lightgreen}\hl{> examine steam}} \\
    \sethlcolor{lightgreen}\hl{A substance called steam. Task completed.}
\end{tcolorbox}
\begin{center}
    Figure 3: Successful sequence of events in the 1-1 Task, where the agent needs to boil water.
\end{center}

\end{document}